\documentclass[10pt,twocolumn,letterpaper]{article}

\usepackage{iccv}
\usepackage{times}
\usepackage{epsfig}
\usepackage{graphicx}
\usepackage{amsmath}
\usepackage{amssymb}
\usepackage[ruled,linesnumbered]{algorithm2e}
\usepackage{algpseudocode}
\usepackage{listings}

\usepackage{float}
\usepackage{graphicx}
\usepackage{multirow}
\usepackage{amsmath} % assumes amsmath package installed
\usepackage{amssymb}  % assumes amsmath package installed
\usepackage[normalem]{ulem}
\usepackage{booktabs}
\usepackage{footnote}
\usepackage{threeparttable}
\usepackage{cite}
\usepackage{color}
\usepackage{booktabs}
\usepackage[table]{xcolor}
\usepackage[pagebackref=true,breaklinks=true,letterpaper=true,colorlinks,linkcolor=red,bookmarks=false]{hyperref}
\pdfoutput=1
\iccvfinalcopy % *** Uncomment this line for the final submission

\ificcvfinal\fi

\begin{document}

\title{GraphAlign: Enhancing Accurate Feature Alignment by Graph matching for Multi-Modal 3D Object Detection}

\author{Ziying Song$^{1}$, Haiyue Wei$^2$, Lin Bai$^1$, Lei Yang$^3$, Caiyan Jia$^{1}$\thanks{Corresponding author}\\
$^1$ School of Computer and Information Technology, Beijing Jiaotong University\\ 
$^2$ School of Information Science and Engineering, Hebei University of Science and Technology\\
$^3$ State Key Laboratory of Automotive Safety and Energy, Tsinghua University
\\
{\tt\small\{songziying, 22120349, cyjia\}@bjtu.edu.cn }\\
{\tt\small ezio59624@gmail.com yanglei20@mails.tsinghua.edu.cn}
\\
}

\maketitle
% Remove page # from the first page of camera-ready.
\ificcvfinal\thispagestyle{empty}\fi

%%%%%%%%% ABSTRACT
\begin{abstract}
   LiDAR and cameras are complementary sensors for 3D object detection in autonomous driving. However, it is challenging to explore the unnatural interaction between point clouds and  images, and the critical factor is how to conduct feature alignment of heterogeneous modalities. Currently, many methods achieve feature alignment by projection calibration only, without considering the problem of coordinate conversion accuracy errors between sensors, leading to sub-optimal performance. In this paper, we present GraphAlign, a more accurate feature alignment strategy for 3D object detection by graph matching. Specifically, we fuse image features from a semantic segmentation encoder in the image branch and point cloud features from a 3D Sparse CNN in the LiDAR branch. To save computation, we construct the nearest neighbor relationship by calculating Euclidean distance within the subspaces that are divided into the point cloud features.  Through the projection calibration between the image and point cloud, we project the nearest neighbors of point cloud features onto the image features. Then by matching the nearest neighbors with a single point cloud to multiple images, we search for a more appropriate feature alignment. In addition, we provide a self-attention module to enhance the weights of significant relations to fine-tune the feature alignment between heterogeneous modalities. Extensive experiments on nuScenes benchmark demonstrate the effectiveness and efficiency of our GraphAlign. 
   %Notably, due to the more accurate feature alignment, which increased mAP by 2.99\% on KITTI test hard level, our method is remarkably beneficial for long-range object detection.
\end{abstract}

%%%%%%%%% BODY TEXT
\section{Introduction}

\begin{figure}[] %%图1
	\centering  %插入的图片居中表示
	\includegraphics[width=1\linewidth]{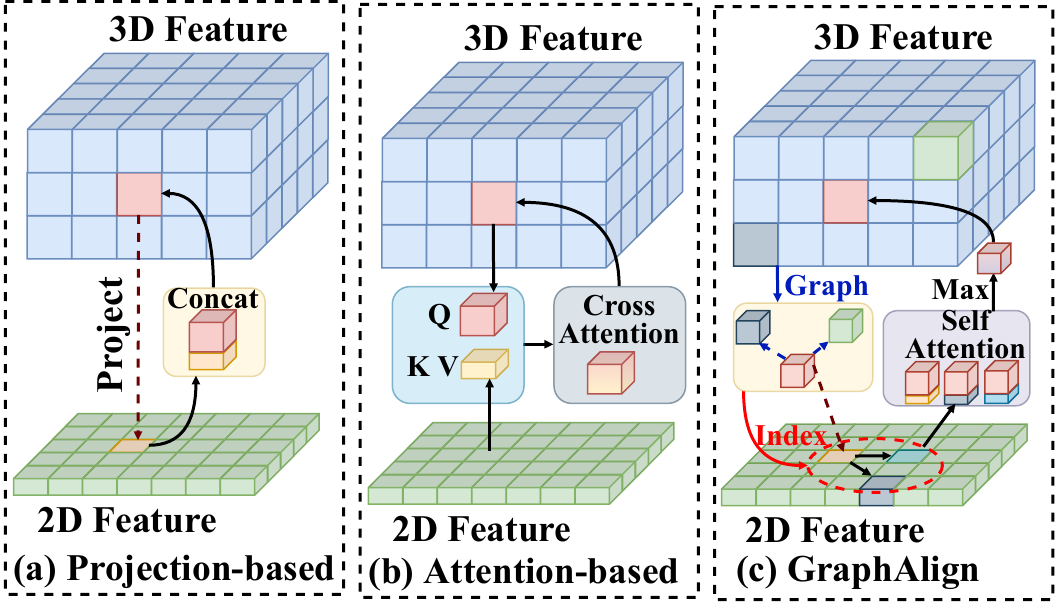}  %插入的图，包括JPG,PNG,PDF,EPS等，放在源文件目录下Hyperparameter
	\caption{Comparison of feature alignment strategies: (a) Projection-based quickly establishes the relationship between modal features but may suffer from misalignment due to sensor error. (b) Attention-based preserves semantic information by learning alignment but has a high computational cost. (c) Our proposed GraphAlign uses graph-based feature alignment to match more plausible alignments between modalities with reduced computation and improved accuracy.
 % Process of different feature alignment strategies. (a) Projection-based quickly constructs the relationship between two modal features, but the inherent accuracy error of the sensor may cause misalignment. (b) Attention-based can better preserve the semantic information in the image by learning alignment rather than projection. However, the computational complexity of attention-based is quadratic with respect to the image size, which may result in significant computation cost. (c) Graph-based feature alignment: We propose to build neighborhoods using graphs to match more plausible alignments between two heterogeneous modalities. Our GraphAlign greatly reduces computation while ensuring more accurate feature alignment.
}  %图片的名称
	\label{fig:motivation}  
\end{figure}
{3}D object detection, a vital computer vision task in autonomous driving, relies on deep learning for accurately identifying and locating objects \cite{moreau2023imposing,xie2021panet,xie2022deep,dai2023eaainet,xie2023deepmatcher,sarlin2023orienternet,song2022fast,song2022msfyolo} in 3D space \cite{shi3d,wang2023multi,wangsong,mao20233d,zou2023object}. With the availability of diverse sensor data, such as cameras and LiDAR, 3D object detection research has made significant progress. However, challenges and difficulties remain due to the inherent limitations of each modality. While LiDAR point cloud provides accurate depth information, it lacks semantic information. Conversely, camera images contain semantic information but lack depth information \cite{shi3d,wangsong}. Therefore, multi-modal 3D object detection has been proposed to leverage the complementary advantages of both modalities to improve detection performance.

Despite the potential of multi-modal 3D object detection, the effective fusion of heterogeneous modal features has not been fully explored. In this work, we mainly attribute the current difficulties of training multi-modal detectors to two aspects. On the one hand, many methods \cite{pointpainting,pointaugmenting,epnet,mvx-net,contfuse,VMVS,pircnn,3d-cvf,seg-voxelnet,mvp,fusionpainting,cat-det,uvtr,focalconv,vpfnet,MV3D,AVOD,SCANet,cross-modality,sfd,graphr-cnn,frustum-pointnets,frustum-convnet,faraway-frustum,frustum-pointpillars,bevfusion-mit,bevfusion-pku} rely on establishing deterministic correspondences between points and image pixels to fuse point clouds and image, as shown in Fig. \ref{fig:motivation} (a). However, accuracy errors resulting from the difference between LiDAR and camera sensors, such as timing synchronization errors, especially the misalignment of small objects in long-range feature fusion, can lead to a decrease in detection performance. On the other hand, a few methods \cite{deepfusion,autoalign,autoalignv2,transfusion} employ attention-based solution to accomplish feature alignment rather than projection, as shown in Fig. \ref{fig:motivation} (b). However, the key issue with using attention-based for point cloud and image feature alignment in multi-modal 3D object detection is that it is too computationally expensive and cannot meet the real-time detection requirements.

In this work, we propose GraphAlign, a graph matching-based feature alignment strategy, to enhance the accuracy of multi-modal 3D object detection, as shown in Fig. \ref{fig:motivation} (c). GraphAlign comprises two key modules: Graph Feature Alignment (GFA) and Self-Attention Feature Alignment (SAFA). The GFA module divides the point cloud space into subspaces and generates the K nearest neighbor features for each point cloud. It then transforms the local neighborhood information of the point cloud into image neighborhood information via a projection calibration matrix, followed by one-to-many feature fusion between a single point cloud feature and K neighbor image features. The SAFA module employs a self-attention mechanism to enhance the weights of important relationships in the fused features and selects the most critical feature from K fused features. Our work's main contributions can be summarized as follows:

 \begin{itemize}
\item We propose GraphAlign, a feature alignment framework based on graph matching, to address the misalignment issue in multi-modal 3D object detection.

\item We propose Graph Feature Alignment (GFA) and Self-Attention Feature Alignment (SAFA) modules to achieve accurate alignment of image features and point cloud features, which can further enhance the feature alignment between point cloud and image modalities, leading to improved detection accuracy.

\item Experiments are conducted using the KITTI\cite{kitti} and nuScenes \cite{nuscenes} benchmarks, demonstrating that GraphAlign can boost point cloud detection accuracy,  especially for long-range object detection.
\end{itemize}

\begin{figure*}[t] %%图2
	\centering  %插入的图片居中表示
	\includegraphics[width=\linewidth]{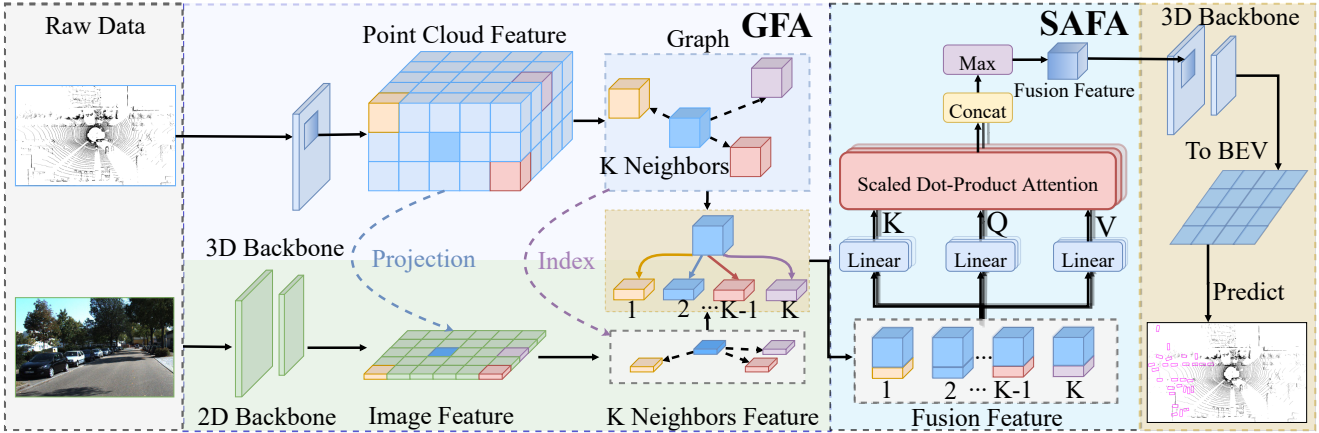}  %插入的图，包括JPG,PNG,PDF,EPS等，放在源文件目录下
	\caption{The framework of GraphAlign. It consists of the Graph Feature Alignment (GFA) module and the Self-Attention Feature Alignment (SAFA) module. The GFA module takes image and point cloud features as input, uses projection calibration matrix to convert 3D positions to 2D pixel positions, constructs local neighborhood information to find nearest neighbors, and combines image and point cloud features. The SAFA module models the contextual relationships among K nearest neighbors through self-attention mechanism, thereby enhancing the importance of fused features, ultimately selecting the most representative features.
 }  %图片的名称
	\label{fig:graphalign}  
\end{figure*}

\section{Related work}
\subsection{3D Object Detection with Single Modality}

3D object detection is commonly conducted using a single modality, either a camera or a LiDAR sensor. Camera-based 3D detection methods\cite{liu2020smoke,bevheight,li2023bevdepth,li2022bevformer,mix-teaching,MonoDIS,yang2023lite,rtm3d,groomed,m3dssd} take images as input and output object localization in space. Some methods use a modified 2D object detection framework with a monocular camera to directly regress 3D box parameters from images\cite{MonoDIS,rtm3d,groomed,m3dssd}. However, monocular cameras cannot provide depth information, which has led to other methods that use stereo or multi-view images to generate dense 3D geometric representations for 3D object detection\cite{imvoxelnet,detr3d}. Although camera-based 3D object detection has made remarkable advancements, its accuracy is not as good as 3D detection methods using LiDAR.

LiDAR-based 3D object detection\cite{xie2023farp,voxelrcnn,second,pointpillars,pointrcnn,Poly-PC} directly processes irregular point cloud data using methods such as PointNet\cite{pointnet} and PointNet++\cite{pointnet++}. Other methods convert point cloud data into regular grids using voxels\cite{voxelnet} and pillars\cite{pointpillars}, which is convenient for feature extraction using 3D or 2D CNN processing\cite{second,voxset,satgcn,vpnet}. Although LiDAR-based 3D object detection is superior to image-based methods, it has limitations due to the sparse nature of point clouds, the lack of texture features, and semantic information.

\subsection{3D Object Detection with Multi-modalities}

To address the limitations of each modality, various methods combine the data from the two modalities to improve detection performance. PointPainting \cite{pointpainting} proposes to enhance each LiDAR point with the semantic score of the corresponding camera image. PI-RCNN\cite{pircnn} fuse semantic features from the image branch and raw LiDAR point clouds to achieve better performance. Frustum PointNets\cite{frustum-pointnets} and Frustum-ConvNet\cite{frustum-convnet} utilize images to generate 2D proposals and then lift them up to 3D space (frustum) to narrow the searching space in point clouds. The Mvx-Net\cite{mvx-net} method appends RoI pooling image eigenvectors to dense eigenvectors for each voxel in a LiDAR point cloud. 3D-CVF\cite{3d-cvf} and EPNet\cite{epnet} explore alignment strategies on feature maps across different modalities with a learned calibration matrix. However, these methods use projection matrices to align two heterogeneous features, which destroys the image semantic information, affecting performance. Other methods propose a learnable alignment method\cite{deepfusion,autoalign,autoalignv2, transfusion} using the cross-attention mechanism. Although this method effectively preserves the semantic information of the image, the frequent query of image features by the attention mechanism increases computational costs.

\section{GraphAlign} \label{sec3}
In this section, we propose an accurate feature alignment, GraphAlign,  which achieves the fusion of point clouds and images by graph matching. We adopt the LiDAR-only detector Voxel RCNN \cite{voxelrcnn} and CenterPoint\cite{centerpoint} as the baseline. Fig. \ref{fig:graphalign} shows the network architecture of our GraphAlign, which includes two modules: Graph Feature Alignment (GFA) module and Self-Attention Feature Alignment (SAFA) module. The details of GraphAlign are presented in the following.

\subsection{  Graph Feature Alignment} \label{subGFA}
Previous works on point cloud and image feature alignment have used projection and attention mechanisms, but these solutions have potential problems. Projection is limited by sensor errors, while attention mechanisms require massive computation. To address these issues, we proposed the Graph Feature Alignment (GFA) module, which constructs neighborhood through graph for more accurate and efficient feature alignment.

The GFA module includes both point cloud and image pipelines, where the fusion of deep features occurs before the 3D Sparse CNN process. Voxel-wise encodor is used to obtain the point cloud features after voxelization. 
We use the depth features of semantic segmenter DeepLabv3 \cite{DeepLabV3} instead of segmentation scores which contain richer appearance cues and larger perception fields, making them more complementary to point cloud fusion. In the projection stage, we treat point clouds as multi-modal aggregation points because point cloud have depth features more suitable for 3D detection than images. We then project the 3D point cloud onto the image plane, as follow:

\begin{equation}\label{equ3d2d}
z_{c}\left[\begin{array}{c}
u \\
v \\
1
\end{array}\right]=h K\left[\begin{array}{ll}
R & T
\end{array}\right]\left[\begin{array}{c}
P_{x} \\
P_{y} \\
P_{z} \\
1
\end{array}\right]
\end{equation}
where, $P_{x}$, $P_{y}$, $P_{z}$ denote the LiDAR point's 3D location, $u$, $v$, $z_{c}$ denote the 2D location and the depth of its projection on the image plane, $K$ denotes the camera intrinsic parameter, $R$ and $T$ denote the rotation and the translation of the LiDAR with respect to the camera reference system, and $h$ denotes the scale factor due to down-sampling.

\begin{figure}[t] %%图3
	\centering  %插入的图片居中表示
	\includegraphics[width=1\linewidth]{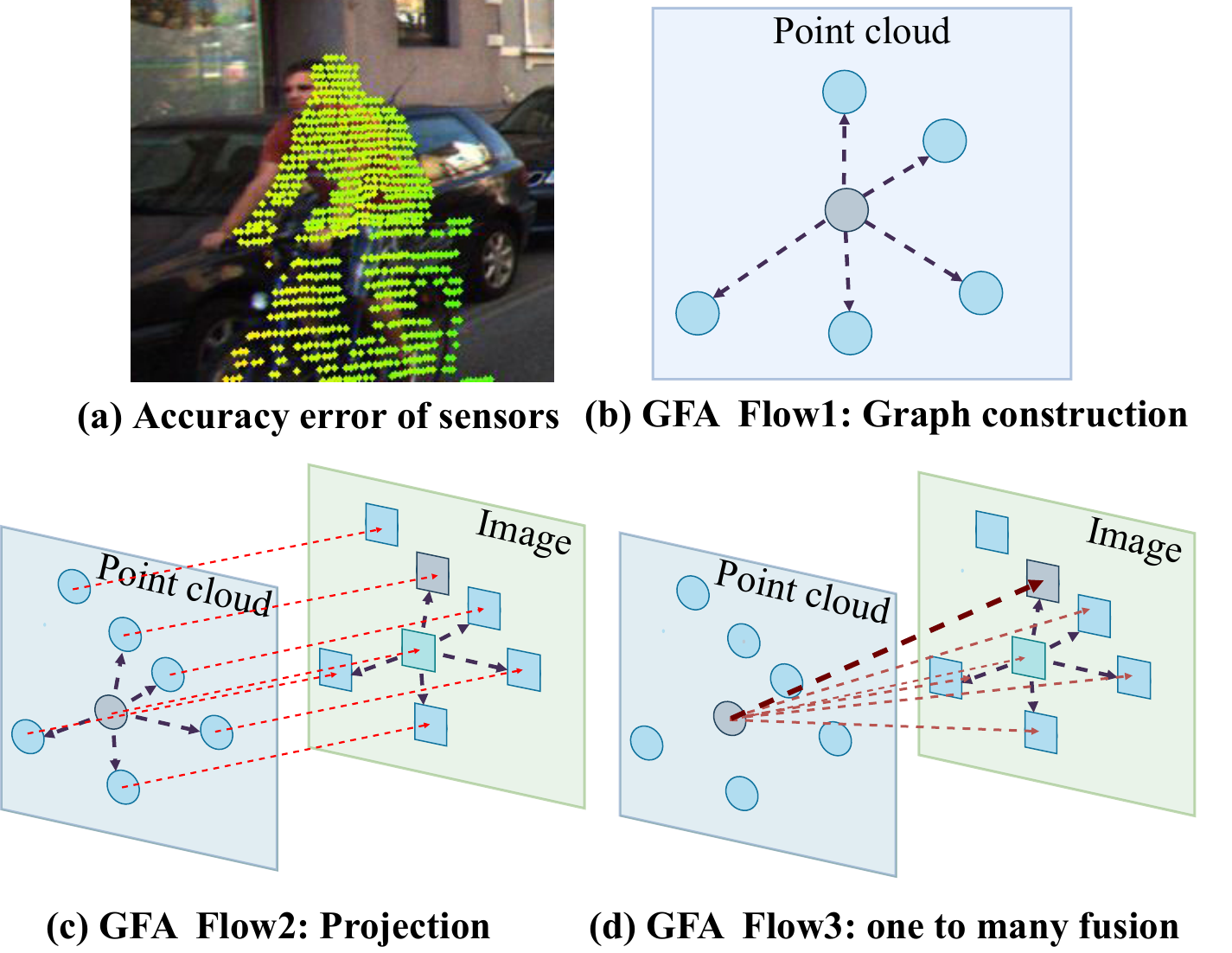}  %插入的图，包括JPG,PNG,PDF,EPS等，放在源文件目录下
	\caption{GFA Process Flow. (a) sensor accuracy errors lead to misalignment. (b) GFA builds neighbor relationships through graphs in the point cloud feature. (c) We project the point cloud features onto the image features and obtain the K nearest neighbors of the image features. (d) We perform one-to-many fusion, specifically, by fusing each individual point cloud feature with K neighboring image features to achieve a better alignment.
 }  %图片的名称
	\label{fig:GFA}  
\end{figure}

After feature extraction, we obtain the point cloud depth feature, defined as $\mathbf{F_{P}} \in \mathbb{R}^{N \times C }$, where $N$, $C$ are the number of point cloud, and channel of the global feature map, respectively. And the 3D coordinates of the point cloud are defined as $\mathbf{C_{P}} \in \mathbb{R}^{N \times 3 }$ , where 3 is the coordinates of the point cloud, represents $(x,y,z)$. To eliminate the impact of feature misalignment due to point cloud data augmentation before 3D to 2D projection, the point cloud is converted to its raw coordinates by inverse operations, such as removing the flip up and down. $\mathbf{C_{P}} $ transforms image $\mathbf{I} \in \mathbb{R}^{h \times w \times 3}$ into pixel coordinates, defined as $\mathbf{C_{I}} \in \mathbb{R}^{N \times  2}$, where 2 is the coordinates of the image pixels, represents $(x,y)$, after the projection calibration matrix Equation (\ref{equ3d2d}). However, since the point cloud is projected onto the image, there exists a small range of image coordinates and an extensive range of point cloud coordinates. We have to remove the coordinates of the image pixels that are out of range after the projection, and the correction rule is $N{}' =\left \{ n_{1}, n_{2}, \dots,n_{j}  \right |  0\le x\le w , 0\le y\le h  \} $, and $N{}'\le N$. Thus, we obtain the filtered novel pixel coordinates, defined as  $\mathbf{C{}'_{I}} \in \mathbb{R}^{N{}' \times  2}$.

To obtain the neighborhood information of the point cloud, defined as  $\mathbf{K_{P}} \in \mathbb{R}^{N \times K }$, where $N$ and $K$ are the number of point cloud and the number of point cloud neighbors, respectively, we performed a data flow in Algorithm (\ref{algorithm:KNN}) and obtained KP by inputting $\mathbf{C_{P}}$ and some hyperparameters. In addition, to save computational cost during this period, we designed subspaces to accelerate the computation, i.e., searching for nearest neighbors in the subspace instead of the whole space.

\begin{algorithm}[t]
\SetAlgoLined
\caption{Graph for the point cloud neighbors} \label{algorithm:KNN}
\KwIn{

Point cloud coordinates  $\mathbf{C_{P}} \in \mathbb{R}^{N \times 3 }$.

Hyper-parameters: No. of point cloud neighbors $K=36$.

Hyper-parameters: No. of point clouds in the subspace $N_{P_{sub}}=5000$.
}

\While{training}{
    No. of subspaces $ N_{sub}= N  {\div} N_{P_{sub}} $

    List: $\mathbf{K_{P}}$
    
    \For{$i_{sub}=1 \dots  N_{sub} $}{
        $ \mathbf{C_{P}}^{i_{sub}}  = [ P_{(i_{sub}-1)\times N_{sub}} ,\dots, P_{N_{sub} \times i_{sub}} ] $
        
        \eIf {$\mathbf{C_{P}}^{i_{sub}} > K $}{
            $D_{sub}$ =$[ \sqrt{( c_{p}^2-c_{p_{i}}^2 )}|c_{p} \in \mathbf{C_{P}}^{i_{sub}},i=1,\dots,N_{P_{sub}} ]$
            
            $\mathbf{K_{P_{sub}}}$=Min($D_{sub}$, K)
            
$\mathbf{K_{P}} = \mathbf{{K_{P}}.Append(\mathbf{K_{P_{sub}}})}$
        }{
            $N_{P_{sub}}$=REM(Num(${C_{P}}$))
            
            $D_{sub}$ =$[ \sqrt{( c_{p}^2-c_{p_{i}}^2 )}|c_{p} \in \mathbf{C_{P}}^{i_{sub}},i=1,\dots,N_{P_{sub}} ]$ 

            $\mathbf{K_{P_{sub}}}$=Min($D_{sub}$, $N_{P_{sub}}$) + $[0,...,0]_{K-N_{P_{sub}}}$
            
$\mathbf{K_{P}} = \mathbf{{K_{P}}.Append(\mathbf{K_{P_{sub}}})}$
        }
        
    }
}

\KwOut{Point cloud neighbors $\mathbf{K_{P}} \in \mathbb{R}^{N \times K \times C }$.}
\end{algorithm}

In addition, we map the neighbors of the point cloud to the image neighbors, defined as $\mathbf{K_{I}} \in \mathbb{R}^{N \times K }$, where $\mathbf{K_{I}} = \mathbf{K_{P}} $. 
% There are three main reasons why we have not constructed neighbor coordinates for images in this process but mapped by point clouds:
% \begin{itemize}
% \item The point cloud coordinates are three-dimensional and have richer Euclidean information than images.
% \item 
% The graph is oriented to input objects with irregular and unstructured data structures like point clouds rather than regular data like images.
% \item 
% There is an insurmountable gap between the 3D point cloud coordinates and the 2D image coordinates of two heterogeneous neighborhoods.
% \end{itemize}
There are three main reasons why we have not constructed neighbor coordinates for images in this process but mapped by point clouds:
First, point cloud coordinates provide richer Euclidean information than images because they are three-dimensional. Second, the graph is designed for input objects with irregular and unstructured data structures like point clouds rather than regular data like images. Third, there is an insurmountable gap between the 3D point cloud coordinates and the 2D image coordinates of two heterogeneous neighborhoods, making it difficult to construct neighbor coordinates for images.

For the above reasons, we choose to index the point cloud neighborhood to the image. Image segmentation encoder outputs image depth features, defined as $\mathbf{F_{I}} \in \mathbb{R}^{h \times w \times c{}'}$, which are indexed by $\mathbf{C{}'_{I}} $ to obtain novel image depth feature $\mathbf {F{}'_{I}} \in \mathbb{R}^{N \times C}$ that are consistent with the point cloud coordinates. Then, indexing the image depth features $\mathbf{F_{}'{I}}$ with $\mathbf{K_{I}}$, we obtain the image depth features with neighbors, defined as $\mathbf{F_{K_{I}}} \in \mathbb{R}^{N \times K \times C} $, where $N$, $K$, and $C$ are the number of point cloud,  the number of image neighbors, and channel of the point cloud feature map, respectively. We replicate the point cloud depth feature  $\mathbf{F_{P}}$ by $K$ times, defined as  $\mathbf{F_{K_{P}}} \in \mathbb{R}^{N \times K \times C}$. Instead of fusing point cloud neighbors and image neighbors, we directly perform $K$ replicated point clouds and $K$ image neighbors fusion for the following reasons. Instead of fusing the point cloud neighbors and the image neighbors, we directly perform the fusion of the replicated $K$ times of the point cloud and the image neighbors to find the most appropriate fusion relationship between points and pixels, as shown in Fig. \ref{fig:GFA}. Finally, we obtain the fused features of point clouds and images with neighborhood relations, as follow:
\begin{equation}\label{equfuse}
\mathbf{F_{K_{PI}}}=\mathbf{F_{K_{P}}}+\mathbf{F_{K_{I}}}
\end{equation}
where, $\mathbf{F_{K_{PI}}} \in \mathbb{R}^{N \times K \times C}$ is the fusion feature.  $N$, $K$, and $C$ are the number of point cloud,  the number of image  neighbors, and channel of the point cloud feature map, respectively.

\subsection{Self-Attention Feature Alignment} \label{subSAFA}
\begin{figure}[t] %%图4
	\centering  %插入的图片居中表示
	\includegraphics[width=0.6\linewidth]{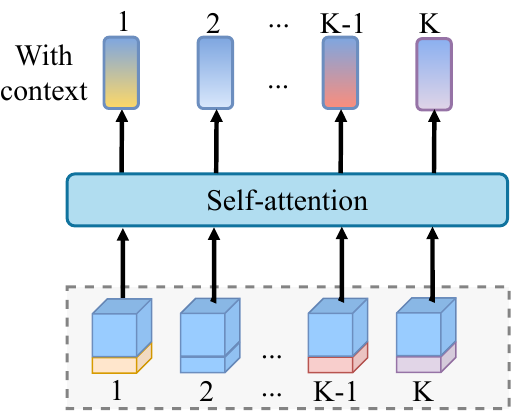}  %插入的图，包括JPG,PNG,PDF,EPS等，放在源文件目录下
	\caption{SAFA module flow. The head and max modules are simplified here, and the SAFA module aims to enhance the expression of fusion features by improving the global context information between the K neighborhoods. % of the point cloud.
 }  %图片的名称
	\label{fig:SAFA}  
\end{figure}

% In the GFA module, we obtain the fusion post-fusion features, which are remarkably effective. However, the aggregation relationship between the fusion features and the K neighbors is underestimated.Therefore, we designed the supplementary Self-Attention Feature Alignment module (SAFA) of GFA to enhance the weights of vital fusion features.
% In the aforementioned GFA module, we obtained the fused features. However, the aggregation relationship between these features and the K neighbors was underestimated. To address this, we designed the SAFA module as a supplement to GFA. The SAFA module uses self-attention to enhance the weights of important fusion features, as shown in Fig. \ref{fig:SAFA}.
After obtaining the fused features in the GFA module as mentioned above, the aggregation relationship between these features and their K neighbors was found to be underestimated. To overcome this limitation, we introduced the SAFA module as a complementary component to the GFA module. It employs self-attention mechanism to amplify the importance of significant fusion features, as illustrated in Fig. \ref{fig:SAFA}.
Currently, a few methods \cite{autoalign,autoalignv2,transfusion,deepfusion} employ cross-attention to learn the feature alignment of point clouds and image heterogeneous modalities. However, the computational cost is high: given the point number $N$ and the size of image feature $W \times H$, the complexity is $O(NWHC)$. Our SAFA module is based on the GFA module and achieves further vital fusion feature weight assignment by performing a multi-head self-attention operation on  $\mathbf{F_{K_{PI}}}$ , and its computation complexity is $O(NK^2C)$, where $N$ is the number of point clouds, $K$ as hyperparameters is mostly 36, $w$ and $h$ are the width and height of the image, which are 1272 and 375 pixels in the KITTI\cite{kitti} dataset, and $C$ is the feature dimension, which is generally 16. By that means, the computation complexity of the cross-attention mechanism-based methods is 400 times higher than our SAFA module.

Specifically, the GFA module outputs fusion features $\mathbf{F_{K_{PI}}} \in \mathbb{R}^{N \times K \times C}$, which we perform a multi-head self-attention operation to find more suitable fusion features with neighborhoods to improve detection performance.
As in Algorithm(\ref{algorithm:SAFA}), we present a remarkably detailed process flow diagram for the SAFA module, which we perform in a calculation-saving manner, and ultimately output a novel fusion feature,defined as $\mathbf{F_{K_{SAFA}}} \in \mathbb{R}^{N \times K \times C}$, where, $N$, $K$, and $C$ are the number of point cloud,  the number of image  neighbors, and channel of the point cloud feature map, respectively. 

The SAFA module aims to enhance the representation of fusion features by better learning the context global information among the K neighborhoods.  Finally, we select the max operation to make up  the more significant features for get a novel fusion feature, defined as $\mathbf{F_{m_{PI}}} \in \mathbb{R}^{N \times C}$, as follow:

\begin{equation}\label{equmax}
\mathbf{F_{m_{PI}}}= Max(\mathbf{F_{K_{SAFA}}})
\end{equation}
where,  $N$, $C$ are the number of point cloud, channel of the point cloud feature map, respectively.

\begin{algorithm}[t]
\SetAlgoLined
\caption{SAFA module workflow} \label{algorithm:SAFA}
\KwIn{

The GFA module outputs fusion features: $\mathbf{F_{K_{PI}}} \in \mathbb{R}^{N \times K \times C}$

Hyper-parameters: No. of attention heads $H$=1.
}
\KwOut{
A noval fusion feature with attention: $\mathbf{F_{K_{SAFA}}} \in \mathbb{R}^{N \times K \times C}$.
}
\While{training}{

    {$W_{\emph{q}}, W_{\emph{k}}, W_{\emph{v}} \in \mathbb{R}^{C \times C  } $} \\

    $Q, K, V = [W_{\emph{q}}, W_{\emph{k}}, W_{\emph{v}}]\mathbf{F_{K_{PI}}}$\
    
    Add $H$ by Reshape: \quad \quad \quad \quad{$Q, K, V \in \mathbb{R}^{N \times K \times C  } $} $->\mathbb{R}^{N \times H \times K \times \frac{C}{H}   }$\\

    $Att = Softmax(QK), Att \in \mathbb{R}^{N \times H \times K \times \frac{C}{H}   }$
    
    $\mathbf{F_{K_{SAFA}}} = Att \mathbf{F_{K_{PI}}}$, $\mathbf{F_{K_{SAFA}}} \in \mathbb{R}^{N \times K \times C}$
    }
\end{algorithm}

\subsection{LiDAR Detection} \label{subGFA}
% After finishing the point cloud and image fusion, we feed $\mathbf{F_{m_{PI}}}$ to Voxel RCNN \cite{voxelrcnn} and CenterPoint \cite{centerpoint},  to complete  subsequent detection. Specifically, our fusion process is completed before the 3D CNN of the LiDAR pipeline. As shown in Fig. \ref{fig:graphalign}, the following process undergoes further feature extraction of the 3D CNN Backbone, after which the feature bird's eye view (BEV), and finally, the detection process is completed by going through the detection head. In addition, the detection of the image is completed by the projection calibration matrix.
% After the point cloud and image fusion, $\mathbf{F_{m_{PI}}}$ is fed to Voxel RCNN \cite{voxelrcnn} and CenterPoint \cite{centerpoint} to complete  subsequent detection. Our fusion process is completed prior to the 3D CNN of the LiDAR pipeline, as illustrated in Fig. \ref{fig:graphalign}. Subsequently, the 3D CNN backbone performs further feature extraction, followed by bird's eye view feature generation and detection head processing to complete the detection process. Furthermore, image detection is accomplished through the projection calibration matrix.
Generally, images serve as auxiliary features for point clouds. After the fusion of point cloud and image, $\mathbf{F_{m_{PI}}}$ is fed into the subsequent LiDAR pipeline for further detection. Our fusion process is completed during the 3D backbone of the LiDAR pipeline, as shown in Fig. \ref{fig:graphalign}. Subsequently, we transform 3D features into BEV features and finally predict using the detection head. In addition, the predicted 3D box of the point cloud can be converted into the 2D box of the image using the projection matrix.

\section{Experiments}\label{Experiments}

In this section, we present the details of each dataset and the experimental setup of GraphAlign, and evaluate the performance of 3D object detection on KITTI\cite{kitti} and nuScenes\cite{nuscenes} datasets.

\subsection{Dataset and Evaluation Metrics}\label{sectionIV-A}
\subsubsection{KITTI dataset} 

% KITTI \cite{kitti} provides synced LiDAR point clouds and front-view camera images. There are 7,481 training samples and 7 518 test samples. Models are commonly evaluated using the mean Average Precision (mAP) metric. The mAP is calculated with recall 40 positions (R40). We conduct experiments on the most commonly used car category and use average Precision (AP) with an (IoU) threshold of 0.7 as evaluation metrics. To further compare the results with other methods on the KITTI 3D detection benchmark, we divide the KITTI training dataset into 4:1 for training and validation and report the performance on the KITTI test dataset.
The KITTI dataset \cite{kitti} provides synchronized LiDAR point clouds and front-view camera images, and consists of 7,481 training samples and 7,518 test samples. The standard evaluation metric for object detection is mean Average Precision (mAP), computed using recall at 40 positions (R40). In this work, we evaluate our models on the most commonly used car category using Average Precision (AP) with an Intersection over Union (IoU) threshold of 0.7. To compare our results with other state-of-the-art methods on the KITTI 3D detection benchmark, we split the KITTI training dataset into a 4:1 ratio for training and validation, and report our performance on the KITTI test dataset.

\subsubsection{nuScenes dataset} 

% nuScenes \cite{nuscenes} is a large-scale 3D detection dataset consisting of 700 scenes for training,150 scenes for validation, and 150 scenes for testing. The dataset was collected using six multi-view cameras and 32-channel LiDAR. 360-degree object annotations for 10 object classes were provided. In evaluation, the main metrics are mAP and nuScenes detection score (NDS), which are produced to quantify a method's detection quality in terms of average classification accuracy, box location, size, orientation, attributes, and velocity.
The nuScenes dataset \cite{nuscenes} is a large-scale 3D detection benchmark consisting of 700 training scenes, 150 validation scenes, and 150 testing scenes. The data were collected using six multi-view cameras and a 32-channel LiDAR sensor, and the dataset includes 360-degree object annotations for 10 object classes. To evaluate the detection performance, the primary metrics used are the mean Average Precision (mAP) and the nuScenes detection score (NDS), which assess a method's detection accuracy in terms of classification, bounding box location, size, orientation, attributes, and velocity. 

% %表1 修改之后修改之后修改之后修改之后修改之后修改之后修改之后
\begin{table}[t]
% \small% 设置字体大小命令由小到大依次为 \tiny \scriptsize \footnotesize \small
% \footnotesize
\scriptsize
\centering
\addtolength{\tabcolsep}{1.3pt}
\caption{Performance comparison with the state-of-the-art methods on KITTI test set for car 3D detection with an average precision of 40 sampling recall points evaluated on KITTI server. ‘L’ and ‘C’ represent LiDAR and Camera, respectively.}
\renewcommand\arraystretch{1}
\setlength{\tabcolsep}{1.4mm}{

\begin{tabular}{c|c|ccc|ccc}
\toprule
\multirow{2}{*}{Method} & \multirow{2}{*}{Modality} & \multicolumn{3}{c|}{AP$_{3D} (\%)$}                                                             & \multicolumn{3}{c}{AP$_{BEV}(\%)$}                                                             \\  \cmidrule(r){3-8} 
                        &                           & \multicolumn{1}{c|}{Easy}           & \multicolumn{1}{c|}{Mod.}           & Hard           & \multicolumn{1}{c|}{Easy}           & \multicolumn{1}{c|}{Mod.}           & Hard           \\ 
                        \midrule

PV-RCNN \cite{pvrcnn}                & L                         & \multicolumn{1}{c|}{90.25}          & \multicolumn{1}{c|}{81.43}          & 76.82          & \multicolumn{1}{c|}{94.98}          & \multicolumn{1}{c|}{90.65}          & 86.14          \\
SECOND \cite{second}                 & L                         & \multicolumn{1}{c|}{84.65}          & \multicolumn{1}{c|}{75.96}          & 68.71          & \multicolumn{1}{c|}{88.07}          & \multicolumn{1}{c|}{79.37}          & 77.50          \\
PointPillars \cite{pointpillars}           & L                         & \multicolumn{1}{c|}{82.58}          & \multicolumn{1}{c|}{74.31}          & 68.99          & \multicolumn{1}{c|}{90.07}          & \multicolumn{1}{c|}{86.56}          & 82.81          \\

VoxSet \cite{voxset}                 & L                         & \multicolumn{1}{c|}{88.53}          & \multicolumn{1}{c|}{82.06}          & 77.46          & \multicolumn{1}{c|}{-}              & \multicolumn{1}{c|}{-}              & -              \\
TANet \cite{tanet}                  & L                         & \multicolumn{1}{c|}{84.39}          & \multicolumn{1}{c|}{75.94}          & 68.82          & \multicolumn{1}{c|}{91.58}          & \multicolumn{1}{c|}{86.54}          & 81.19          \\
Part-A$^{2}$ \cite{parta2}                & L                         & \multicolumn{1}{c|}{85.94}          & \multicolumn{1}{c|}{77.86}          & 72.00          & \multicolumn{1}{c|}{89.52}          & \multicolumn{1}{c|}{84.76}          & 81.47          \\
VP-Net \cite{vpnet}                & L                         & \multicolumn{1}{c|}{90.46}          & \multicolumn{1}{c|}{82.03}          & 79.65          & \multicolumn{1}{c|}{94.49}          & \multicolumn{1}{c|}{90.99}          & 86.58          \\

MV3D \cite{MV3D}                   & L\&C                      & \multicolumn{1}{c|}{74.97}          & \multicolumn{1}{c|}{63.63}          & 54.00          & \multicolumn{1}{c|}{86.62}          & \multicolumn{1}{c|}{78.93}          & 69.80          \\

MMF \cite{mmf}                    & L\&C                      & \multicolumn{1}{c|}{86.81}          & \multicolumn{1}{c|}{76.75}          & 68.41          & \multicolumn{1}{c|}{89.49}          & \multicolumn{1}{c|}{87.47}          & 79.10          \\

PI-RCNN \cite{pircnn}                & L\&C                      & \multicolumn{1}{c|}{84.37}          & \multicolumn{1}{c|}{74.82}          & 70.03          & \multicolumn{1}{c|}{91.44}          & \multicolumn{1}{c|}{85.81}          & 81.00          \\
EPNet \cite{epnet}                  & L\&C                      & \multicolumn{1}{c|}{89.81}          & \multicolumn{1}{c|}{79.28}          & 74.59          & \multicolumn{1}{c|} {94.22}          & \multicolumn{1}{c|}{88.47}          & 83.69          \\
PointPainting \cite{pointpainting}          & L\&C                      & \multicolumn{1}{c|}{82.11}          & \multicolumn{1}{c|}{71.70}          & 67.08          & \multicolumn{1}{c|}{-}              & \multicolumn{1}{c|}{-}              & -              \\
MSF-MC \cite{msf-mc}                 & L\&C                      & \multicolumn{1}{c|}{89.63}          & \multicolumn{1}{c|}{80.06}          & 75.83          & \multicolumn{1}{c|}{93.42}          & \multicolumn{1}{c|}{86.97}          & 84.54          \\
Fast-CLOCs \cite{fastclocs}             & L\&C                      & \multicolumn{1}{c|}{89.11}          & \multicolumn{1}{c|}{80.34}          & 76.98          & \multicolumn{1}{c|}{93.02}          & \multicolumn{1}{c|}{89.49}          & 86.39          \\
Focals Conv \cite{focalconv}            & L\&C                      & \multicolumn{1}{c|}{90.55}          & \multicolumn{1}{c|}{82.28}          & 77.59          & \multicolumn{1}{c|}{-}              & \multicolumn{1}{c|}{-}              & -              \\
SFD \cite{sfd}           & L\&C                                   & \multicolumn{1}{c|}{91.73}          & \multicolumn{1}{c|}{\textbf{84.76}}          & \multicolumn{1}{c|}{77.92}            & \multicolumn{1}{c|}{95.64}   & \multicolumn{1}{c|}{\textbf{91.85}}    & 86.83         \\
HMFI  \cite{hmfi}           & L\&C                   & \multicolumn{1}{c|}{88.90}          & \multicolumn{1}{c|}{81.93}          & \multicolumn{1}{c|}{77.30}          & \multicolumn{1}{c|}{-}   & \multicolumn{1}{c|}{-}      & -        \\
Graph-VoI \cite{graphr-cnn}              & L\&C                                   & \multicolumn{1}{c|}{\textbf{91.89}}          & \multicolumn{1}{c|}{83.27}          & \multicolumn{1}{c|}{77.78}            & \multicolumn{1}{c|}{\textbf{95.69}}   & \multicolumn{1}{c|}{90.10}    & 86.85         \\
\midrule
Voxcl RCNN \cite{voxelrcnn}           & L                         & \multicolumn{1}{c|}{90.90}          & \multicolumn{1}{c|}{81.62}          & 77.06          & \multicolumn{1}{c|}{-}              & \multicolumn{1}{c|}{-}              & -              \\
Voxcl RCNN*             & L                         & \multicolumn{1}{c|}{90.76}          & \multicolumn{1}{c|}{81.69}          & 77.42          & \multicolumn{1}{c|}{92.89}          & \multicolumn{1}{c|}{89.97}          & 84.69          \\
\rowcolor{blue!10} Our GraphAlign          & L\&C                      & \multicolumn{1}{c|}{90.96} & \multicolumn{1}{c|}{83.49} & \textbf{80.14} & \multicolumn{1}{c|}{93.91}          & \multicolumn{1}{c|}{91.79} & \textbf{88.05} \\ 
\bottomrule
\end{tabular}
}
\label{tab_kitti_test}
\begin{tablenotes}
\footnotesize
\item[1] * denotes re-implement result.
\end{tablenotes}
\end{table}

% \begin{table}[htp]
% % \small% 设置字体大小命令由小到大依次为 \tiny \scriptsize \footnotesize \small
% % \footnotesize
% \scriptsize
% \centering
% \addtolength{\tabcolsep}{1.3pt}
% \caption{Comparison with SOTA methods on nuScenes test set.  }
% \renewcommand\arraystretch{1}
% \setlength{\tabcolsep}{3.8mm}{
% \begin{tabular}{c|c|c|c}
% \toprule
% Method                & mAP (\%) & NDS(\%) & FPS (A6000) \\ \midrule
% AutoAlignV2           & 68.4 & 72.4 &4.8\\
% BEVFusion (ICRA)            & 70.2 & 72.9 &4.2\\
% % BEVFusion (NeurIPS)            & 69.2 & 71.8 &-\\(没有找到时间)
% TransFusion            & 68.9 & 71.7 &5.5\\
% DeepInteraction-base & \textbf{70.8} & \textbf{73.4}&3.1 \\ \midrule
% \rowcolor{blue!10} GraphAlign        & 66.5 & 70.6& \textcolor{black}{\textbf{27.0}} \\ \bottomrule
% \end{tabular}}
% \label{tab_nus_test_rebuttal}
% \end{table}
\subsection{ Implementation Details}

\subsubsection{Network Architecture}
% Since KITTI\cite{kitti} and nuScenes\cite{nuscenes} are different datasets, we describe the settings of GraphAlign for each dataset in the following section. 

% \textbf{GraphAlign with Voxel RCNN\cite{voxelrcnn}}  We validate our GraphAlign on KITTI\cite{kitti}, which adopts baseline is Voxel RCNN \cite{voxelrcnn}. Where the input voxel size is set as (0.05m, 0.05m, 0.1m), the anchor size of cars is [3.9, 1.6, 1.56], and anchor rotations are [0, 1.57]. We adopt the same data augmentation solution as Focals Conv\cite{focalconv}.

% \textbf{GraphAlign with CenterPoint\cite{centerpoint}} We validate our GraphAlign on nuScenes\cite{nuscenes}, which adopts baseline is CenterPoint\cite{centerpoint}. Where we set the detection range to [-54m, 54m] for the X and Y axis and [-5m, 3m] for the Z axis. And the input voxel size is set as (0.075m, 0.075m, 0.2m), and the maximum number of point clouds contained in each voxel is 10.
Since KITTI \cite{kitti} and nuScenes \cite{nuscenes} are distinct datasets with varying evaluation metrics and characteristics, we provide a detailed description of the GraphAlign settings for each dataset in the following section.

\textbf{GraphAlign with Voxel RCNN \cite{voxelrcnn}:} We validate our GraphAlign on the KITTI \cite{kitti} dataset using Voxel RCNN \cite{voxelrcnn} as the baseline. The input voxel size is set to (0.05m, 0.05m, 0.1m), with anchor sizes for cars set at [3.9, 1.6, 1.56], and anchor rotations at [0, 1.57]. We adopt the same data augmentation solution as Focal Loss \cite{focalconv}.

\textbf{GraphAlign with CenterPoint \cite{centerpoint}:} We validate our GraphAlign on the nuScenes \cite{nuscenes} dataset using CenterPoint \cite{centerpoint} as the baseline. The detection range for the X and Y axis is set at [-54m, 54m] and [-5m, 3m] for the Z axis. The input voxel size is set at (0.075m, 0.075m, 0.2m), and the maximum number of point clouds contained in each voxel is set to 10.

%表2  修改之后修改之后修改之后修改之后修改之后修改之后修改之后
\begin{table}[t]
\scriptsize
\centering
\caption{Performance comparison with state-of-the-art methods on KITTI validation dataset for car class. The results are reported by the mAP with 0.7 IoU threshold and 40 recall points. ‘L’ and ‘C’ represent LiDAR and Camera, respectively.}
\renewcommand\arraystretch{1}
\setlength{\tabcolsep}{1.3mm}{

\begin{tabular}{@{}@{\extracolsep{\fill}}!{\color{white}\vline}l|c|c|c|c|c|c|c @{}}
\toprule
\multirow{2}{*}{Method} & \multirow{2}{*}{Moiality} & \multicolumn{3}{c|}{AP$_{3D} (\%)$}   & \multicolumn{3}{c}{AP$_{BEV} (\%)$}                                                                                                                      \\ \cmidrule(r){3-8} 
                        &                           & Easy           & Mod.           & Hard    & Easy           & Mod.           & Hard            \\ \midrule
PointRCNN \cite{pointrcnn}              & L                         & 88.88          & 78.63          & 77.38    & \_             & \_             & \_           \\
% H$^2$3D R-CNN \cite{deng2021multi}              & L                         & 89.63          & 85.20          & 79.08    & \_             & \_             & \_           \\

SECOND \cite{second}                 & L                         & 87.43         & 76.48          & 69.10   & \_             & \_             & \_                \\
% MedTr-TSD \cite{tian2023medoidsformer}                 & L                         &  89.27        &  84.24          & 78.85   & \_             & \_             & \_                \\
CT3D \cite{ct3d}                   & L                         & \textbf{92.85} & 85.82          & 83.46    & 96.14	& 91.88	 & 89.63
          \\

Part-A$^{2}$ \cite{parta2}                & L                         & 89.47          & 79.47          & 78.54    & 90.42          & 88.61          & 87.31               \\
MV3D \cite{MV3D}                   & L\&C                      & 71.29          & 62.68          & 56.56    & 86.55 	& 78.10 	& 76.67 
           \\

MMF \cite{mmf}                    & L\&C                      & 87.90          & 77.87          & 75.57       & \textbf{96.66}             & 88.25             & 79.60         \\
MSF-MC \cite{msf-mc}                 & L\&C                      & \multicolumn{1}{c|}{89.63}          & \multicolumn{1}{c|}{80.06}          & 75.83          & \multicolumn{1}{c|}{93.42}          & \multicolumn{1}{c|}{86.97}          & 84.54      \\
PI-RCNN \cite{pircnn}                & L\&C                      & 88.27          & 78.53          & 77.75     & \_             & \_             & \_             \\
EPNet \cite{epnet}                  & L\&C                      & 92.28          & 82.59          & 80.14     & 95.51 	& 91.47 	& 91.16 \\
\midrule        
Voxel RCNN \cite{voxelrcnn}             & L                         & 92.38          & 85.29          & 82.86   & 95.52          & 91.25          & 88.99            \\
\rowcolor{blue!10} Our GraphAlign          & L\&C                      & \multicolumn{1}{c|}{92.44}          & \multicolumn{1}{c|}{\textbf{87.01}} & \textbf{84.68}& 95.65  & \textbf{92.82}   & \textbf{91.41}   \\ 
\bottomrule
\end{tabular}
}
\label{tab_kitti_val}
\end{table}

\subsubsection{Training and Testing Details}
% Our GraphAlign is trained from the beginning via Adam optimizer and adopts a standalone semantic segmenter DeepLabv3\cite{DeepLabV3}. For training KITTI\cite{kitti} and nuScenes\cite{nuscenes}, we use GPUs with 8 NVIDIA RTX A6000 to train the whole network. For KITTI, our GraphAlign (Voxel RCNN \cite{voxelrcnn})  takes about 2 hours. For nuScenes, our GraphAlign (CenterPoint \cite{centerpoint})  takes about 20 hours. In model inference stage, we first perform non-maximal suppression (NMS) in RPN with an IoU threshold of 0.7 and keep the top-100 region proposals as the input of detection head. Then, after refinement, NMS is applied again with an IoU threshold of 0.1 to remove redundant predictions. Please refer to OpenPCDet \cite{openpcdet} for more detailed configuration since we conduct all experiments with this toolkit.

Our GraphAlign is meticulously trained from scratch using the Adam optimizer and incorporates a stand-alone semantic segmenter, namely, DeepLabv3 \cite{DeepLabV3}. To enable effective training on KITTI \cite{kitti} and nuScenes \cite{nuscenes}, we utilize 8 NVIDIA RTX A6000 GPUs for network training. Specifically, for KITTI, our GraphAlign model based on Voxel RCNN \cite{voxelrcnn} requires approximately 2 hours of training time which train 80 epochs. Whereas for nuScenes, our GraphAlign model based on CenterPoint \cite{centerpoint} necessitates approximately 20 hours of training time which train 20epochs. During the model inference stage, we employ a non-maximal suppression (NMS) operation in RPN with an IoU threshold of 0.7 and select the top 100 region proposals to serve as input for the detection head. Following refinement, we apply NMS again with an IoU threshold of 0.1 to eliminate redundant predictions. For additional details concerning our method, please refer to OpenPCDet \cite{openpcdet}.
% Our GraphAlign is trained using the Adam optimizer and DeepLabv3\cite{DeepLabV3} as a semantic segmenter. Training is performed on 8 NVIDIA RTX A6000 GPUs for KITTI\cite{kitti} and nuScenes\cite{nuscenes}. For KITTI, GraphAlign (Voxel RCNN \cite{voxelrcnn}) takes approximately 2 hours, and for nuScenes, GraphAlign (CenterPoint \cite{centerpoint}) takes approximately 20 hours. During inference, non-maximal suppression (NMS) is applied with an IoU threshold of 0.7 in RPN, and the top-100 region proposals are input to the detection head. After refinement, NMS is applied again with an IoU threshold of 0.1 to remove redundant predictions. Please refer to OpenPCDet \cite{openpcdet} for detailed configuration, as all experiments are conducted using this toolkit.
\subsection{Comparison with State-of-the-Arts}

% In this section, we perform extensive experiments on KITTI\cite{kitti} and nuScenes\cite{nuscenes}  datasets comparison with the state-of-the-art methods, achieving extremely advanced results.
%表3   修改之后修改之后修改之后修改之后修改之后修改之后修改之后
\begin{table*}[t]
\scriptsize
\centering
\caption{ Comparison with SOTA methods on the nuScenes test set. ‘C.V.’, ‘Ped.’, and ‘T.C.’ are short for construction vehicle, pedestrian, and traffic cone, respectively. }
\renewcommand\arraystretch{1}
\setlength{\tabcolsep}{2.6mm}{

% \begin{tabular}{cccccccccccccc}
\begin{tabular}{c|c|c|c|c|c|c|c|c|c|c|c|c}
\toprule
Method             & mAP  & NDS  & Car  & Truck & C.V. & Bus  & Trailer & Barrier & Motor. & Bike & Ped. & T.C. \\ 
\midrule
InfoFocus  \cite{infofocus}               & 39.5 & 39.5 & 77.9 & 31.4  & 10.7 & 44.8 & 37.3    & 47.8    & 29.0   & 6.1  & 63.4 & 46.5 \\
S2M2-SSD  \cite{s2m2}              & 62.9 & 69.3 & 86.3 & 56.0  & 26.2 & 65.4 & 59.8    & 75.1    & 61.6   & 36.4  & 84.6 & 77.7 \\
AFDetV2  \cite{afdetv2}            & 62.4 & 68.5 & 86.3 & 54.2  & 26.7 & 62.5 & 58.9    & 71.0    & 63.8   & 34.3  & 85.8 & 80.1 \\
VISTA  \cite{vista}            & 63.0 & 69.8 & 84.4 & 55.1  & 25.1 & 63.7 & 54.2    & 71.4    & 70.0   & 45.4  & 82.8 & 78.5 \\
PointPillars\cite{pointpillars} & 30.5& 45.3 &68.4 &23.0& 4.1 &28.2 &23.4 &38.9 &27.4& 1.1& 59.7& 30.8\\
PointPainting \cite{pointpainting}       & 46.4 & 58.1 & 77.9 & 35.8  & 15.8 & 36.2 & 37.3    & 60.2    & 41.5   & 24.1 & 73.3 & 62.4 \\
MVP  \cite{mvp}                 & 66.4 & 70.5 & 86.8 & 58.5  & 26.1 & 67.4 & 57.3    & 74.8    & 70.0   & 49.3 & 89.1 & 85.0 \\
AutoAlign\cite{autoalign} & 65.8 & 70.9 & 85.9 & 55.3 & 29.6 & 67.7 & 55.6 & - & 71.5 & 51.5 & 86.4 & -\\
AutoAlignV2\cite{autoalignv2} & 68.4 & 72.4 & 87.0 & 59.0 & 33.1 & 69.3 & 59.3 & - & 72.9 & 52.1 & 87.6 & -\\
TransFusion \cite{transfusion} & 68.9 & 71.7 & 87.1 & 60.0 & 33.1 & 68.3 & 60.8 & 78.1 & 73.6 & 52.9 & 88.4 & 86.7\\
BEVFusion \cite{bevfusion-pku} & 69.2 & 71.8 & 88.1 & 60.9 & 34.4 & 69.3 & 62.1 & 78.2 & 72.2 & 52.2 & 89.2 & 85.2\\
DeepInteraction \cite{deepinteraction} &70.8& 73.4& 87.9& 60.2& 37.5& 70.8& 63.8& 80.4& 75.4& 54.5& 90.3& 87.0 \\

\midrule
CenterPoint \cite{centerpoint}                  & 58.0 & 65.5 & 84.6 & 51.0 & 17.5 & 60.2 & 53.2 & 70.9 & 53.7 & 28.7 & 83.4 & 76.7 \\
\rowcolor{blue!10} Our GraphAlign     &  66.5\textit{\fontsize{6}{0}\selectfont\textcolor{red}{+8.5}} &	70.6\textit{\fontsize{6}{0}\selectfont\textcolor{red}{+5.1}} &	87.6&	57.7  &	26.1\textit{\fontsize{6}{0}\selectfont\textcolor{red}{+8.6}}  &	66.2 &	57.8  &	74.1 &	72.5\textit{\fontsize{6}{0}\selectfont\textcolor{red}{+18.8}}  &	49.0\textit{\fontsize{6}{0}\selectfont\textcolor{red}{+20.3}}  &   87.2  &	86.3\textit{\fontsize{6}{0}\selectfont\textcolor{red}{+9.6}} \\ 
\bottomrule
\end{tabular}
}
\label{tab_nuScens_test}
\end{table*}

% 修改之后表4
\begin{table}[]
\scriptsize
\centering
\caption{Effect of each component in our GraphAlign. Results are reported on KITTI validation set with Voxel RCNN. "P" indicates projection.}
\renewcommand\arraystretch{1}
\setlength{\tabcolsep}{2.2mm}{
\begin{tabular}{c|c|c|c|c|c|c}
\toprule
 \multirow{2}{*}{P} & \multirow{2}{*}{GFA} & \multirow{2}{*}{SAFA} & \multicolumn{2}{c|}{AP$_{3D}(\%)$}                                        & \multirow{3}{*}{\#Params} & \multirow{3}{*}{Runtime} \\ \cmidrule(r){4-5}
                                                 &                      &                                  & Mod.           & Hard                     &                                   &                       \\ 
                        \midrule
                                      &                      &                       & 85.29 & 82.86          & 7.59 M                            & 5ms                   \\
              \checkmark                         &                      &                       & 85.59\textit{\fontsize{6}{0}\selectfont\textcolor{red}{+0.30}} & 83.07\textit{\fontsize{6}{0}\selectfont\textcolor{red}{+0.21}}       & 7.74 M                           & 17ms                  \\
  \checkmark                         & \checkmark                     &                                  & 86.62\textit{\fontsize{6}{0}\selectfont\textcolor{red}{+1.03}}  &  84.21\textit{\fontsize{6}{0}\selectfont\textcolor{red}{+1.14}}  & 7.74 M                           & 24ms                  \\
\rowcolor{blue!10}     \checkmark                        & \checkmark                     & \checkmark                                & 87.01\textit{\fontsize{6}{0}\selectfont\textcolor{red}{+0.39}}  &  84.68\textit{\fontsize{6}{0}\selectfont\textcolor{red}{+0.47}}         & 7.75 M                           & 26ms                  \\ 
     \bottomrule
\end{tabular}
}
\label{tab_GFA_SAFA_ablation_kitti}
\end{table}

% 修改之后表5

% \begin{table}[]
% \scriptsize
% \centering
% \caption{Improvement over multi-modal baseline on nuScenes $\frac{1}{4} $. "P" indicates projection}
% \renewcommand\arraystretch{1}
% \setlength{\tabcolsep}{1.2mm}{
% \begin{tabular}{c|c|c|c|c|c|c|c}
% \toprule
% Method                       & P & GFA & SAFA & mAP  & NDS  & \#Params & Runtime \\ \midrule
% \multirow{4}{*}{CenterPoint\cite{centerpoint}} &         &     &      & 56.1 & 64.2 & 9.01M    & 12ms    \\
%                              &    \checkmark      &     &      & 58.0 & 64.8 & 9.16M    & 28ms    \\
%                              &     \checkmark     & \checkmark     &      & 61.3 & 67.8 & 9.16M    & 35ms    \\
%                              &      \checkmark    &  \checkmark    &   \checkmark    & 62.8 & 68.5 & 9.17M    & 37ms   \\ \bottomrule
%                              }
% \end{tabular}\label{tab_GFA_SAFA_ablation_nuscens}
% \end{table}
% 加上了 autoalignV2 一行

\subsubsection{Performance on KITTI dataset.}
As shown in Table \ref{tab_kitti_test} , we compare GraphAlign with state-of-the-art methods in 3D and BEV APs on KITTI test dataset. We observe that our GraphAlign achieves state-of-the-art performance. In detail, it shows remarkably outstanding performance at three difficulty levels of 3D and BEV detection, with (90.96\%, 83.49\%, 80.14\%, 93.91\%, 91.79\%, 88.05\%). For better comparison, we reproduce Voxel RCNN \cite{voxelrcnn} as a strong baseline network. It is worth noting that our replication is almost identical to the results reported in \cite{voxelrcnn}. Our GraphAlign achieves better performance than the baseline Voxel RCNN with (0.06\%, 1.80\%, 2.72\%, 1.02\%, 1.82\%, 3.36\%) improvements on three levels. Compared with the multi-modal method Focals Conv \cite{focalconv}, our GraphAlign achieves better performance than Focals Conv with the improvements  (0.41\%, 1.21\%, 2.72\%).  Our GraphAlign performs well on the KITTI \cite{kitti} test dataset's moderate and hard levels, which have more long-range objects. In addition, we also provide the results of the KITTI validation dataset to better present the detection performance of our GraphAlign, as shown in Table \ref{tab_kitti_val}. There is a significant improvement compared to the baseline Voxel RCNN on the KITTI \cite{kitti} validation dataset's moderate and hard levels. Even slight misalignments between point clouds and images can likely lead to significant errors in detection. GraphAlign's ability to use graphs to establish relationships between heterogeneous modalities is a key factor in its success.

\subsubsection{Performance on nuScenes dataset.}
% As shown in Table \ref{tab_nuScens_test}nuScenes,  we compare with evaluate our models on the test server and compare them with Image-only, LIDAR-only, and multi-modal methods, as show in Table \ref{tab_nuScens_test}. The proposed method surpasses previous single-modal methods with 65.4\% mAP and 69.6\% NDS, outperforming the strong CenterPoint baseline by 7.4\% mAP and 4.1\% NDS. As for ambiguous classes like motorcycle and bicycle, the gain is even up to 17.8\% AP. It also surpasses the recently developed multi-modal 3D detector PointPainting\cite{pointpainting} and Fast-CLOCs\cite{fastclocs}.
% mvp autoalignv2 transfusion pointaugmenting. slightly weak
% Although the mAP and NDS of our GraphAlign are slightly weaker than MVP, AutoAlignV2, Transfusion, PointAugmenting. 
% we are the first to propose using graphs to achieve heterogeneous modal feature alignment for 3D multi-modal detectors.
We also conducted experiments on the much larger nuScenes \cite{nuscenes} dataset using the state-of-the-art 3D detector CenterPoint \cite{centerpoint} to further validate the effectiveness of our GraphAlign. As shown in Table \ref{tab_nuScens_test}, GraphAlign achieved 66.5 mAP and 70.6 NDS on the nuScenes test dataset, which is 8.5 mAP and 5.1 NDS higher than the strong CenterPoint\cite{centerpoint}. Furthermore, we marked in red those categories, "C.V.," "Motor," "Bike," and "T.C.," that showed a significant increase in performance, with "Motor" and "Bike" increasing by 18.8\% and 20.3\%, respectively. Our GraphAlign exhibits excellent performance in small object detection, particularly for these four categories that are characterized by a higher proportion of small objects at long distances, following more accurate feature alignment.

% 表五
\begin{table}[t]
\scriptsize
\centering
\caption{Effect of each component in our GraphAlign. Results are reported on nuScenes $\frac{1}{4} $ validation with CenterPoint. "P" indicates projection.
}
\renewcommand\arraystretch{1}
\setlength{\tabcolsep}{2.3mm}{
\begin{tabular}{c|c|c|c|c|c|c}
\toprule
P & GFA & SAFA & mAP  & NDS  & \#Params & Runtime \\ \midrule
        &     &      & 56.1 & 64.2 & 9.01M    & 12ms    \\
                                 \checkmark      &     &      & 58.0\textit{\fontsize{6}{0}\selectfont\textcolor{red}{+1.9}}  & 64.8\textit{\fontsize{6}{0}\selectfont\textcolor{red}{+0.6}}  & 9.16M    & 28ms    \\
                                  \checkmark     & \checkmark     &      & 61.3\textit{\fontsize{6}{0}\selectfont\textcolor{red}{+3.3}}  & 67.8\textit{\fontsize{6}{0}\selectfont\textcolor{red}{+3.0}}  & 9.16M    & 35ms    \\
                                \rowcolor{blue!10}   \checkmark    &  \checkmark    &   \checkmark    & 62.8\textit{\fontsize{6}{0}\selectfont\textcolor{red}{+1.5}}  & 68.5\textit{\fontsize{6}{0}\selectfont\textcolor{red}{+0.7}}  & 9.17M    & 37ms   \\ 
                                % \cmidrule(r){1-7}
        % & \multicolumn{2}{c|}{AutoAlignV2*}       & 63.0 & 68.1 & 13.21M    & 207ms \\
\bottomrule                             
\end{tabular}
}
\label{tab_GFA_SAFA_ablation_nuscens}
\begin{tablenotes}
\footnotesize
\item[1] * denotes re-implement result.
\end{tablenotes}
\end{table}

\subsection{ Ablation Study}\label{subsection_Ablation}

\subsubsection{Effect of Projection-only, the GFA and SAFA modules, and Attention-based}
This section discusses the results of ablation experiments conducted on the baseline detectors Voxel RCNN and CenterPoint to evaluate the performance of each module in GraphAlign. The results are reported in Table \ref{tab_GFA_SAFA_ablation_kitti} and Table \ref{tab_GFA_SAFA_ablation_nuscens} for KITTI and nuScenes $\frac{1}{4}$ validation datasets, respectively.

As shown in Table \ref{tab_GFA_SAFA_ablation_kitti}, the moderate and hard AP scores for KITTI are initially at 85.29\% and 82.86\%, respectively. Adding the projection-only module to the image branch only slightly improves the mAP score by 0.30\% and 0.21\%, respectively. The lackluster performance was due to the intrinsic misalignment caused by the sensors' accuracy errors. However, compared with projection-only, the consecutive addition of GFA and SAFA modules led to a continuous improvement of AP on the moderate and hard levels of KITTI, by 1.03\% and 0.39\%  (moderate) and 1.14\% and 0.47\% (hard), respectively. This significant performance improvement is attributed to our GraphAlign method's accurate alignment of long-range objects.

As shown in Table \ref{tab_GFA_SAFA_ablation_nuscens}, the Attention-based feature alignment strategy AutoAlignV2\cite{autoalignv2}  shows slightly improved detection accuracy compared to GraphAlign, its runtime is five times longer and cannot meet the real-time requirement. The reason for this is the global alignment learning between the point cloud and the image through the Attention mechanism, while our GraphAlign learns local relationships based on projection, similar to the Anchor mechanism, to enable faster learning based on prior knowledge. Although the SAFA module with an attention mechanism has been added to our GraphAlign, it focuses on learning important local features of neighbors rather than computing with global image features.

\subsubsection{Effect of the Hyperparameters}
In this section, we have analyzed the experimental results of our model with respect to various hyperparameters, including $K$ (number of point cloud neighbors), $N_{P_{sub}}$ (number of point clouds in the subspace), and $H$ (number of attention heads) for the car class on the KITTI validation dataset and $\frac{1}{4}$ on nuScenes.

Table \ref{tab_K} indicates that the optimal performance for GraphAlign based on Voxel RCNN was achieved when $K$ was set to 16, while for GraphAlign based on CenterPoint, the optimal value was 25. Although there was a slight improvement in performance on easy and moderate levels for $K$ set to 25 and 36, respectively, the corresponding increase in computation time by 11% and 30% was taken into account.

Moreover, Table \ref{tab_N} reveals that the runtime was significantly impacted by $N_{P_{sub}}$. We found that the best performance was achieved with $N_{P_{sub}}$ set to 1000. Even though setting $N_{P_{sub}}$ to 3000 resulted in a slight improvement in accuracy on the KITTI dataset, the increase in AP was limited, and the runtime increased by 38\%. Therefore, we selected 1000 as the optimal value for $N_{P_{sub}}$. Subdividing the subspace proved critical in saving time, and setting $N_{P_{sub}}$ to 500 did not improve performance.

Additionally, as depicted in Fig. \ref{tab_H}, as $H$ increases, AP gradually improves, but the effect on runtime is more significant, with an increase of approximately 10\%. Generally, $H=1$ is preferred, but larger values may be used to achieve higher AP.

% 表6  修改之后修改之后修改之后修改之后修改之后修改之后修改之后
\begin{table}[t]
\scriptsize
\centering
\caption{Effect of the number of point cloud neighbors $K$.}
\renewcommand\arraystretch{1}
\setlength{\tabcolsep}{2.2mm}{
\begin{tabular}{c|c|c|c|c|c|c|c}
\toprule
\multirow{4}{*}{$K$} & \multicolumn{4}{c|}{KITTI \cite{kitti}}                                                               & \multicolumn{3}{c}{nuScenes \cite{nuscenes}}                                           \\ \cmidrule(r){2-8} 
                   & \multicolumn{3}{c|}{AP$_{3D}$(\%)}  & \multirow{3}{*}{Runtime} & \multirow{3}{*}{mAP} & \multirow{3}{*}{NDS} & \multirow{3}{*}{Runtime} \\ \cmidrule(r){2-4}
                   & Easy     & Mod.    & Hard      &                          &                      &                      &                          \\ \midrule 
 9	& 91.53	& 86.38	& 84.17	& 25ms	& 61.9	& 67.9	& 35ms  \\ 
  16	& 92.44	& 87.01	& \textbf{84.68}	& 26ms	&
62.1	& 67.8	& 37ms   \\
25	& 92.09	& \textbf{87.11}	& 84.23	& 29ms	& 62.8	& \textbf{68.5}	& 41ms  \\ 
 36	& \textbf{92.58}	& 86.87	& 83.97	& 34ms	& 62.9	& 68.3	& 47ms  \\ 
48	& 92.43	&87.58	& 83.90 &	40ms &	\textbf{63.1} &	68.1 &	54ms  \\ 
\bottomrule
\end{tabular}
}\label{tab_K}
\end{table}

\begin{table}[t]
\scriptsize
\centering
\caption{Effect of the number of point cloud $N_{P_{sub}}$ in the subspace.}
\renewcommand\arraystretch{1}
\setlength{\tabcolsep}{1.9mm}{
% Please add the following required packages to your document preamble:
% \usepackage{multirow}
\begin{tabular}{c|c|c|c|c|c|c|c}
\toprule
\multirow{4}{*}{$N_{P_{sub}}$} & \multicolumn{4}{c|}{KITTI \cite{kitti}}                                                               & \multicolumn{3}{c}{nuScenes \cite{nuscenes}}                                           \\ \cmidrule(r){2-8} 
                   & \multicolumn{3}{c|}{AP$_{3D}$(\%)} & \multirow{3}{*}{Runtime} & \multirow{3}{*}{mAP} & \multirow{3}{*}{NDS} & \multirow{3}{*}{Runtime} \\ \cmidrule(r){2-4}
                   & Easy     & Mod.    & Hard        &                          &                      &                      &                          \\ \midrule 
                   
500               & 92.09    & 86.59   & 84.01    & 23ms                     & 60.9                 & \textbf{68.8}                & 34ms                     \\
1000               & 92.44    & 87.01   & 84.68     & 26ms                     & \textbf{62.8}                 & 68.5                & 37ms                     \\
3000               & \textbf{92.57}    & \textbf{87.05}   & 84.77    & 36ms                     & 61.9                & 68.7                 & 50ms                     \\
5000               & 92.09    & 86.81   & 84.12     & 45ms                     & 62.2                 & 67.9                 & 58ms                     \\
8000               & 91.97    & 86.24   & \textbf{85.02}      & 53ms                     & 61.8                 & 67.5                 & 69ms                     \\
10000              & 92.11    & 86.27   & 84.32      & 59ms                     & 61.5                 & 67.0                 & 76ms                     \\ \bottomrule
\end{tabular}
}
\label{tab_N}
\end{table}

\begin{table}[t]
\scriptsize
\centering
\caption{Effect of the number of attention heads $H$.}
\renewcommand\arraystretch{1}
\setlength{\tabcolsep}{2.2mm}{
\begin{tabular}{c|c|c|c|c|c|c|c}
\toprule
\multirow{4}{*}{$H$} & \multicolumn{4}{c|}{KITTI \cite{kitti}}                                                               & \multicolumn{3}{c}{nuScenes \cite{nuscenes}}                                           \\ \cmidrule(r){2-8} 
                   & \multicolumn{3}{c|}{AP$_{3D}$(\%)}  & \multirow{3}{*}{Runtime} & \multirow{3}{*}{mAP} & \multirow{3}{*}{NDS} & \multirow{3}{*}{Runtime} \\ \cmidrule(r){2-4}
                   & Easy     & Mod.    & Hard      &                          &                      &                      &                          \\ \midrule 
1	& 92.44	& 87.01	& 84.68	& 26ms	&  62.8	& 68.5	& 37ms   \\
2	& \textbf{93.57}	& 87.89	& 84.79	& 29ms	& 62.9	& 68.4	& 41ms   \\
3	& 92.87	& 87.69	& 84.55	& 35ms	&  63.1	& 68.4	& 47ms  \\
4	& 93.51	& \textbf{88.13}	& \textbf{84.97}	& 41ms	& \textbf{63.5}	& \textbf{68.8}	& 56ms  \\ 
\bottomrule
\end{tabular}
}\label{tab_H}
\end{table}

\begin{table}[]
\scriptsize
\centering
\caption{Performance on different distances. The results are evaluated with 3D AP calculated by 40 recall positions for car
class on the moderate level.}
\renewcommand\arraystretch{1}
\setlength{\tabcolsep}{0.8mm}{
\begin{tabular}{c|c|c|c|c|c|c|c|c}
\toprule
 \multirow{3}{*}{P}     & \multirow{3}{*}{GFA} & \multirow{3}{*}{SAFA} & \multicolumn{3}{c|}{AP$_{3D}(\%)$}                                                            & \multicolumn{3}{c}{AP$_{BEV}(\%)$}                                      \\ \cmidrule(r){4-9} 
                            &                        &                                             & 0-20m & 20-40m & 40m-inf & 0-20m & 20-40m & 40m-inf \\ \midrule
\multirow{4}{*}  &                      &                       & 95.94 & 79.45 & 39.45   & 96.09 & 89.34  & 54.00   \\
                             \checkmark &                      &                       & 95.97 & 79.52\textit{\fontsize{6}{0}\selectfont\textcolor{red}{+0.07}}  & 38.93\textit{\fontsize{6}{0}\selectfont\textcolor{red}{-0.52}}   & 96.48 & 90.68\textit{\fontsize{6}{0}\selectfont\textcolor{red}{+1.34}}  & 53.54\textit{\fontsize{6}{0}\selectfont\textcolor{red}{-0.46}}   \\
                             \checkmark & \checkmark                    &                       & 96.13 & 82.02\textit{\fontsize{6}{0}\selectfont\textcolor{red}{+2.50}}  & 46.49\textit{\fontsize{6}{0}\selectfont\textcolor{red}{+7.56}}   & 96.47 & 92.65\textit{\fontsize{6}{0}\selectfont\textcolor{red}{+1.88}}  & 58.56\textit{\fontsize{6}{0}\selectfont\textcolor{red}{+5.20}}   \\
\rowcolor{blue!10}                             \checkmark & \checkmark                    & \checkmark                     & 96.12 & 82.17\textit{\fontsize{6}{0}\selectfont\textcolor{red}{+0.15}}  & 47.67\textit{\fontsize{6}{0}\selectfont\textcolor{red}{+1.18}}   & 96.57 & 93.54\textit{\fontsize{6}{0}\selectfont\textcolor{red}{+0.89}}  & 59.99\textit{\fontsize{6}{0}\selectfont\textcolor{red}{+1.43}}   \\ \bottomrule
\end{tabular}
}
\label{tab_distances}
\end{table}

\subsubsection{Distances Analysis}
 To better understand the excellent performance of our GraphAlign at long distances, we present performance metrics for different distance ranges in Table \ref{tab_distances}. Specifically, there was a significant drop in Projection-only, particularly in the 40m-inf range where there were many small objects. This was mainly due to misalignment of small objects from different modalities. By adding the GFA and SAFA modules, the 3D APs increased by 7.56\% and 1.18\% respectively, while the BEV APs increased by 5.20\% and 1.43\% respectively in the 40m-inf range. These results demonstrated that our GraphAlign's effective feature alignment strategy is helpful for the long-range small objects detection.

\section{Conclusions}

In this work,we present GraphAlign, a more accurate and efficient feature alignment strategy for 3D object detection by graph matching. Specifically, the Graph Feature Alignment (GFA) module constructs neighbor fusion features based on prior knowledge of projection matrix, which avoids sensor accuracy errors. Furthermore, the Self-Attention Feature Alignment (SAFA) module is designed to enhance the weight of important relationships. Comprehensive experimental results demonstrate that GraphAlign significantly improves the 3D detector on the KITTI and nuScenes datasets. Building upon existing Projection-based and Attention-based feature alignment strategies, we hope that our work can provide a new perspective for multi-modal feature fusion in autonomous driving.
% In this work, we analyzed the core problem of inefficient and inaccurate perception of distant objects in multi-modal fusion 3D object detection. On one hand, the projection-based methods heavily rely on projection calibration and are constrained by the accuracy errors of the sensors, resulting in poor long-distance detection performance. On the other hand, the attention-based methods, although achieving feature alignment, have low efficiency and cannot meet real-time detection requirements. We propose an efficient feature alignment strategy based on Graph and design a high-efficiency and high-precision 3D object detection framework, GraphAlign, suitable for long distance objects. we provides a novel strategie for feature alignment in multi-modal 3D object detection.

\textbf{Limitation and future work.} One of our limitations is that our GraphAlign rely on independent semantic segmentation rather than end-to-end learning. As such, in future work, we aim to investigate the use of end-to-end learning for graph matching in multi-modal fusion.

\section*{Acknowledgments}
This work was supported by 
%the Fundamental Research Funds for the Central Universities (2023YJS019), 
the National Key R\&D Program of China (2018AAA0100302), and the STI 2030-Major Projects under Grant 2021ZD0201404 (Funded by Jun Xie and Zhepeng Wang from Lenovo Research). We sincerely appreciate the helpful discussions provided by Dr. Shaoqing Xu from University of Macau. 
% In addition, we we extend our heartfelt appreciation to Lenovo Research for providing the supercomputing platform.

{\small
\bibliographystyle{ieee_fullname}
\bibliography{egbib}
}

\end{document}